\documentclass{article}
\usepackage{spconf,amsmath,graphicx,hyperref,booktabs,multirow}

\title{INTERMITTENT SEMI-WORKING MASK: A NEW MASKING PARADIGM FOR LLMS}
\name{
\begin{tabular}{c}
HaoYuan Hu\textsuperscript{1,2,*}\thanks{These authors contributed equally to this work.}, Mingcong Lu\textsuperscript{2,*}, Di Luo\textsuperscript{2}, XinYa Wu\textsuperscript{2}, Jiangcai Zhu\textsuperscript{2}, Taoye Yin\textsuperscript{2}, Zheng Li\textsuperscript{2},  \\
Hao Wang\textsuperscript{2}, Shusheng Zhang\textsuperscript{2}, KeZun Zhang\textsuperscript{2}, KaiLai Shao\textsuperscript{2}, Chao Chen\textsuperscript{2}, Feng Wang\textsuperscript{2,}\sthanks{Corresponding author: zifan.wf@antgroup.com}
\end{tabular}
}

\address{Zhejiang University\textsuperscript{1} \\ Ant Group\textsuperscript{2} }

\begin{document}

\maketitle
\begin{abstract}
%Multi-turn dialogues are a key interaction method between humans and Large Language Models (LLMs), as conversations extend over multiple rounds, keeping LLMs’ high generation quality and low latency is a challenge. Mainstream LLMs can be grouped into two categories based on masking strategy: causal LLM and prefix LLM. Several works have demonstrated that prefix LLMs tend to outperform causal ones in scenarios that heavily depend on historical context such as multi-turn dialogues or in-context learning, thanks to their bidirectional attention on prefix sequences. However, prefix LLMs have an inherent inefficient training problem in multi-turn dialogue datasets. In addition, the attention mechanism of prefix LLM makes it unable to reuse Key-Value Cache (KV Cache) across dialogue rounds to reduce generation latency. In this paper, we propose a novel masking scheme called Intermittent Semi-working Mask (ISM) to address these problems. Specifically, we apply alternate bidirectional and unidirectional attention on queries and answers in the dialogue history. In this way, ISM is able to maintain the high quality of prefix LLM and low generation latency of causal LLM, simultaneously. Extensive experiments illustrate that our ISM achieves significant performance.

Multi-turn dialogues and context-intensive tasks challenge Large Language Models (LLMs) to integrate long histories without sacrificing generation quality.
Although prefix LLMs can better exploit historical context via bidirectional attention on prefix tokens, they are rarely used in practice because multi-turn training requires many duplicated triplets, and its bidirectional prefix prevents KV-cache reuse at inference time, driving up high cost and latency.
To retain the contextual understanding of prefix mask while preserving the inference-time efficiency of causal mask, we introduce Intermittent Semi-working Mask (ISM), a masking scheme that injects sparse bidirectional attention into the causal backbone.
ISM alternates bidirectional attention over query segments with unidirectional attention over answer segments, enabling the synthesis of in-context while preserving global causality.
This design eliminates triplet expansion during training and maintains KV-cache reuse during inference, yielding latency comparable to standard causal LLMs.
ISM is architecture-agnostic and parameter-free, adding only minimal latency.
Across extensive evaluations, ISM outperforms causal baselines not only on multi-turn dialogue, but also on context-intensive tasks like mathematical reasoning.
\end{abstract}

\section{Introduction}
\label{sec:intro}

Large Language Models (LLMs) achieve strong results across many Natural Language Process (NLP) tasks, with multi-turn dialogue as the primary interaction mode; yet, as conversation histories grow, maintaining high answer quality under low latency remains challenging.
Modern LLMs are predominantly decoder-only and use causal mask~\cite{brown2020language, achiam2023gpt}, which enforces unidirectional attention, with each token attending only to preceding tokens, thereby enabling efficient training and KV-cache reuse.
Although causal mask can improve efficiency, empirical evidence~\cite {anil2023palm,tay2022ul2,chung2022scalinginstructionfinetunedlanguagemodels,raffel2020exploring} shows that unidirectional mask is overly restrictive for context-conditioned generation, and other work~\cite{ding2023causallm} further proves that prefix mask is a better choice theoretically.
Prefix mask~\cite{du2022glm} can enhance contextual synthesis via bidirectional attention over prefix tokens. 
In practice, however, bidirectional dependencies prevent KV-cache reuse across dialogue turns at inference time, causing high latency, and multi-turn training requires expanding each dialogue into many duplicated triplets, whereas causal mask covers all turns in a single forward pass.
As a result, mainstream LLMs remain causal mask despite their contextual limitations.
This gap motivates approaches that retain the contextual understanding of prefix mask while preserving the inference-time efficiency of causal mask.
%This gap motivates methods that recover prefix-like contextual integration without sacrificing training efficiency or KV-cache reuse, with benefits extending beyond dialogue to other context-intensive tasks such as mathematical reasoning.

% However, in practice, multi-turn training requires expanding each dialogue into many duplicated triplets-whereas causal mask cover all turns in a single forward pass, while bidirectional dependencies prevent KV-cache reuse across dialogue turns during inference, leading to high latency. 
% As a result, mainstream systems remain causal despite their contextual limitations.

\begin{figure}[tbp]
    \centering
    \includegraphics[width=0.48\textwidth]{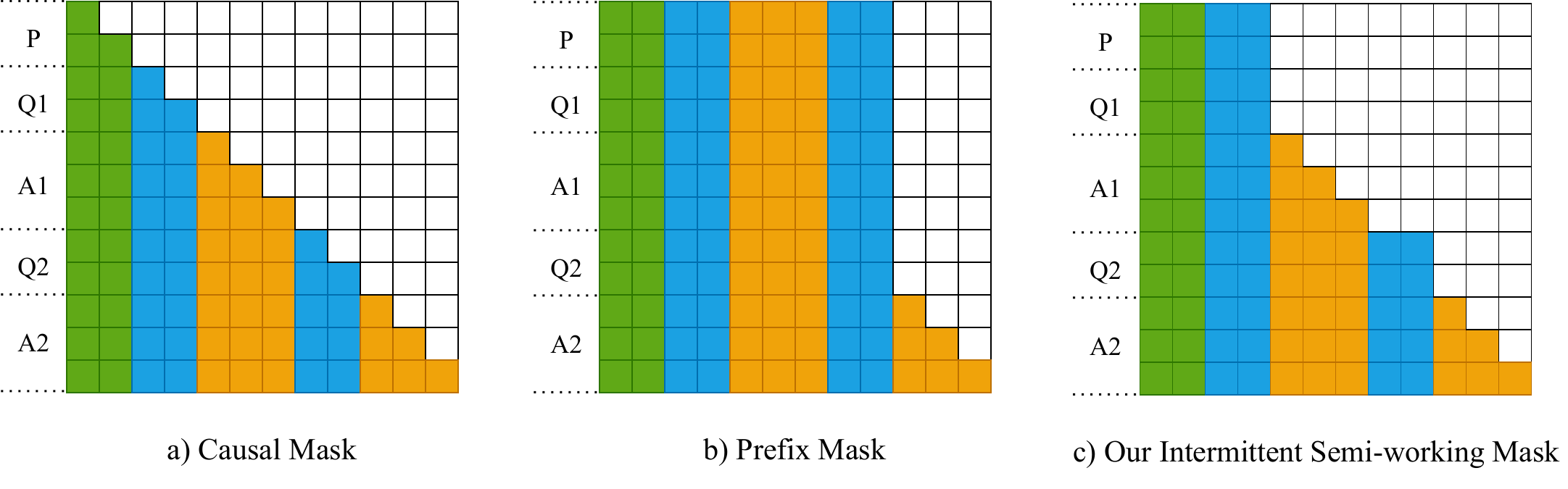}
    \caption{Illustration of our Intermittent Semi-working Mask \textit{vs.} existing Causal Mask and Prefix Mask.} %in Multi-Turn Dialogues. 
    %Taking the second round of dialogue as an example, we show the mask difference between our method and existing works.
    % the color filling part means tokens are attendable
    %The dialogue history (\textbf{P}rompt+\textbf{Q}uery\textbf{1}+\textbf{A}nswer\textbf{1}) and current \textbf{Q}uery\textbf{2} serve as prefix sequences, LLMs should output \textbf{A}nswer\textbf{2}. Causal Mask employs unidirectional attention on prefix sequences, while Prefix Mask applies bidirectional attention. Our ISM utilizes alternate bidirectional and unidirectional attention on queries and answers in prefix sequences. All of them generate answer in auto-regressive.}
    \label{introfig}
\end{figure}

In this paper, we propose Intermittent Semi-working Mask (ISM), a new masking scheme that delivers prefix-mask–level response quality with causal-mask–level latency.
As illustrated in Figure~\ref{introfig}, given a dialogue history and current query $[\mathbf{P}, \mathbf{Q_1}, \mathbf{A_1}, \cdots, \mathbf{Q_{n-1}}, \mathbf{A_{n-1}}, \mathbf{Q_n}]$, ISM applies bidirectional attention within query segments ($\mathbf{Q_i}$) and strictly unidirectional attention within answer segments ($\mathbf{A_i}$), while preventing attention to subsequent blocks. Different from prefix mask, ISM enables full KV-cache reuse across dialogue turns at inference time and avoids expanding multi-turn dialogues into multiple single-turn samples during training. 
ISM is drop-in compatible with mainstream causal LLMs, requires no architectural changes or additional parameters.

We integrate ISM into two decoder-only causal backbones (LLaMA and Qwen) and evaluate on two multi-turn dialogue benchmarks and a math benchmark, where ISM achieves a quality–latency trade-off and delivers notable gains.
Overall, our contribution can be summarized as follows:
\begin{itemize}
    \item We introduce ISM, an intermittent masking scheme that applies bidirectional attention to query segments and unidirectional attention to answer segments, while preventing attention to future blocks, which delivers prefix-mask–level response quality with causal-mask–level inference-time latency.
    \item ISM is drop-in compatible with mainstream causal LLMs, requires no architectural changes or additional parameters, preserves KV-cache reuse across turns, and enables efficient multi-turn training without expanding dialogues into multiple single-turn samples.
    \item Extensive experiments across two famous backbones (LLaMA and Qwen) and three benchmarks demonstrate significant improvements in response quality, while maintaining near-identical end-to-end latency.
    %We deploy our ISM in a real-world online environment to test its practical performance, we observe our ISM brings significant improvements in generation latency and quality.
\end{itemize}

\section{Methodology}
In this section, we first give the mathematical description of ISM. Then, we theoretically explain the advantages of our ISM in terms of convergence. In addition, we also introduce how to integrate ISM and reuse KV-cache practically.

\subsection{Mathematical Description}
Let the input be $\mathbf{X} = (\mathbf{x}_1 ,\cdots, \mathbf{x}_n)$.
%A standard Softmax Self-Attention (SSA) layer updates
%\begin{equation*}
%    \mathbf{x}_j \gets \mathbf{x}_j + \mathbf{P} \mathbf{V} \mathbf{X} \mathrm{softmax}(\mathbf{X}^\top \mathbf{K}^\top \mathbf{Q}  \mathbf{x}_j),
%\end{equation*} 
Let $\mathbf{Q},\mathbf{K},\mathbf{V}$ denote the query, key, and value projections, and $\mathbf{O}$ the output projection (we omit scaling factors, positional encoding, and multi-head notation for clarity).
Following prior work~\cite{von2023transformers, zhang2023trained}, we analyze the Linear Self-Attention (LSA) surrogate:

%Because the softmax nonlinearity complicates analysis
\begin{equation}
        \mathbf{x}_j \gets \mathbf{x}_j + \mathbf{O} \mathbf{V} \sum_{i=1}^n \mathbf{x}_i \left(\mathbf{x}_i^\top \mathbf{K}^\top \mathbf{Q}  \mathbf{x}_j\right).
    \label{e_full}
\end{equation}

We consider a multi-turn dialogue sample $[\mathbf{P}, \mathbf{Q}_1, \mathbf{A}_1,$ $\cdots,\mathbf{Q}_{n-1}, \mathbf{A}_{n-1}, \mathbf{Q}_n]$,
where prompt $\mathbf{P}=(\mathbf{x}_1, \cdots, \mathbf{x}_{m_p})$,
the $k$-th query $\mathbf{Q}_k=(\mathbf{x}_{m_{a_{k-1}}+1},...,\mathbf{x}_{m_{q_k}})$ with $m_{a_0} := m_p$,
and the $k$-th answer $\mathbf{A}_k=(\mathbf{x}_{m_{q_k} +1},\cdots, \mathbf{x}_{m_{a_k}})$ for $k\ge 1$.
The goal is to generate the answer $\mathbf{A}_n$ for query $\mathbf{Q}_n$ given the historical context $[\mathbf{P}, \mathbf{Q}_1, \mathbf{A}_1,\cdots, \mathbf{Q}_{n-1}, \mathbf{A}_{n-1}]$.

\begin{enumerate}
    \item Full (or bidirectional) attention corresponds to Eq.~\ref{e_full}, where every token attends to all positions.
    \item Causal (or unidirectional) attention (standard in transformer decoder) restricts each token to its past:
    \begin{equation}
        \mathbf{x}_j \gets \mathbf{x}_j + \mathbf{O} \mathbf{V} \sum_{i=1}^j \mathbf{x}_i \left(\mathbf{x}_i^\top \mathbf{K}^\top \mathbf{Q}  \mathbf{x}_j\right),
        \label{e_causal}
    \end{equation}
    so token $\mathbf{x}_j$ can attend only to positions $1$ through $j$.
    \item Prefix attention enables full attention up to the end of the current query $\mathbf{Q}_n$ while remaining causal thereafter:
    \begin{equation}
        \mathbf{x}_j \gets \mathbf{x}_j + \mathbf{O} \mathbf{V} \sum_{i=1}^{max(j, m_{q_n})} \mathbf{x}_i \left(\mathbf{x}_i^\top \mathbf{K}^\top \mathbf{Q}  \mathbf{x}_j\right),
        \label{e_prefix}
    \end{equation}
    so token with $(j\leq m_{q_n})$ can attend to the entire prefix.
\end{enumerate}

\begin{figure}
    \centering
    \includegraphics[width=0.48\textwidth]{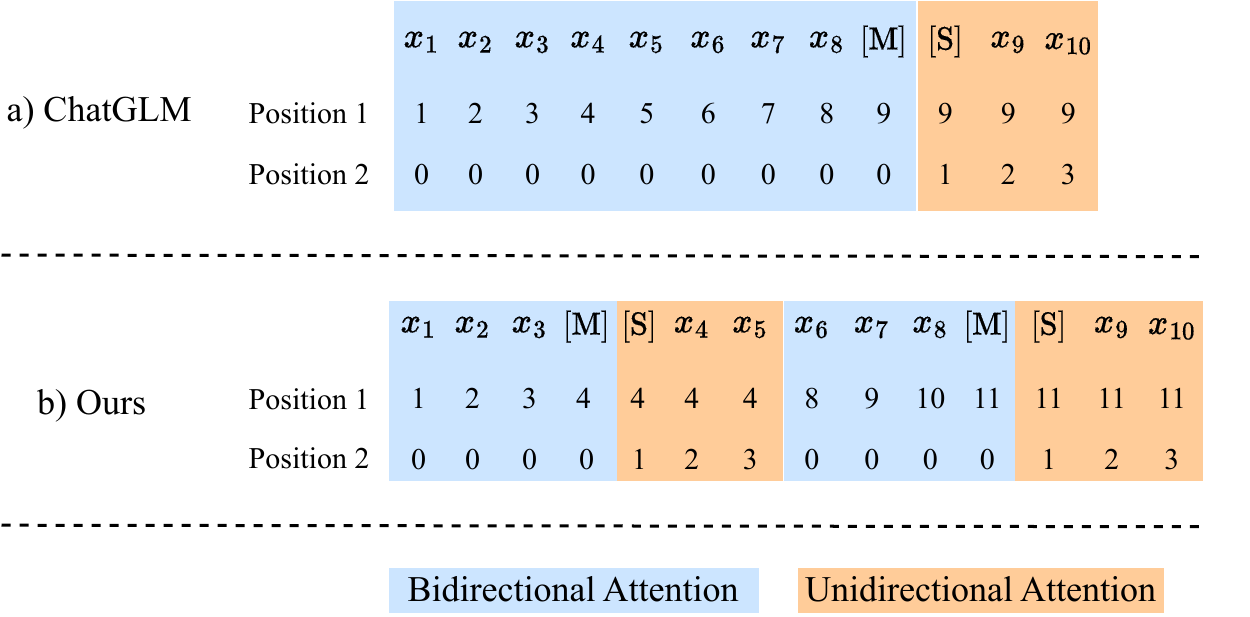}
    \caption{Comparison between ISM and ChatGLM when facing multi-turn dialogue data. [M]:=[MASK], [S]:=[START].} %ChatGLM applies bidirectional attention (blue frame) to prefix tokens (dialogue history) when generating answer (orange frame), our ISM applies alternate bidirectional and unidirectional attention to previous queries and answers when generating an answer. }
    \label{glm_pos_compare}
\end{figure}

Intermittent Semi-working Mask (ISM) inserts bidirectional attention within each query segment while keeping unidirectional attention in answer segments and forbidding attention to future segments:
\begin{equation}
\mathbf{x}_j \gets \mathbf{x}_j + \mathbf{O} \mathbf{V} \sum_{i=1}^{f(j)} \mathbf{x}_i \left(\mathbf{x}_i^\top \mathbf{K}^\top \mathbf{Q} \mathbf{x}_j\right),
\label{e_ISM}
\end{equation}
\begin{equation}
    f(j) = 
    \begin{cases}
        m_{q_1}, &\text{for}\quad 1\le j \le m_{q_1}  \\
        m_{q_{k}}, &\text{for}\quad m_{a_{k-1}}< j \le m_{q_{k}},\quad k > 1 \\
        j, &\text{for}\quad m_{q_{k}}< j \le m_{a_{k}},\quad k >= 1 \\
    \end{cases}.
    \label{fj}
\end{equation}
Thus, tokens within query $\mathbf{Q}_k$ can attend bidirectionally to all tokens up to the end of that query (including all prior context), while tokens in answer $\mathbf{A}_k$ use strictly causal attention. Crucially, no token attends beyond the end of its current query, ensuring previously computed keys and values remain valid in future turns. Hence, KV-caches can be reused across dialogue turns as in causal LLMs, while still enabling contextual integration of prefix LLMs within each query.

\subsection{Convergence Advantage}\label{Convergence}

We adopt the LSA surrogate above and construct $\mathbf{K},\mathbf{Q},\mathbf{V},\mathbf{O}$ so that each layer exactly mirrors one step of gradient descent on a linear regression problem.
Under this construction, the residual updates reduce to sums over the attention-accessible positions. As proofed in prior work~\cite{ding2023causallm}:
\begin{enumerate}
    \item Prefix mask grants every prefix token access to all in-context examples, yielding the same dynamics as batch gradient descent and thus linear convergence of the iterative weights $\mathbf{w}^{(l)}$ to the least-squares optimum $\mathbf{w}^*$.
    \item Causal mask limits each position $j$ to its past, inducing position-dependent online gradient descent. The iterates $\mathbf{w}_j^{(l)}$ converge to stationary points $\mathbf{w}_j^*$ that are generally suboptimal and need not approach $\mathbf{w}^*$ when the number of in-context examples is finite.
    \item ISM replaces the causal bound $j$ with a segment-wise boundary $f(j)$. For tokens in the final query window $(m_{n-1}<j\le m_n)$, the updates aggregate all in-context examples, recovering batch gradient descent dynamics and linear convergence to $\mathbf{w}^*$ for the query. 
\end{enumerate}
 
Therefore, ISM retains the optimal in-context convergence of prefix mask while avoiding the global inefficiencies of pure bidirectional attention at inference time.

\subsection{Practical Integration}\label{practical_modify}

Building on our theoretical result that ISM converges to the optimal stationary point $\mathbf{w}^*$ in typical settings ($m_n > m_{n-1}$), we implement ISM by modifying the standard causal mask (Figure~\ref{introfig}a) to insert bidirectional attention over query spans, yielding the ISM pattern (Figure~\ref{introfig}c).

A key practical advantage of ISM is KV-cache reuse across dialogue turns, which substantially lowers generation latency. Given a history $h =[\mathbf{P},\mathbf{Q}_1,\mathbf{A}_1,\cdots,\mathbf{Q}_{n-1},\mathbf{A}_{n-1}]$ and current query $\mathbf{Q}_n$, both causal mask and ISM reuse the cached keys/values for the history ($C_h$), compute the cache for the current query ($C_q$), and concatenate $[C_h; C_q]$ to generate $A_n$ with linear complexity. In contrast, prefix mask must recompute $C_h$ when processing $\mathbf{Q}_n$ because bidirectional dependencies span the entire prefix, preventing cache reuse.

We also demonstrate ISM on a prefix-style backbone by adapting ChatGLM~\cite{du2022glm}.
ChatGLM employs 2D positional encoding (Position 1 encodes the span identity and Position 2 encodes the intra-span offset) and bidirectional attention over prefix tokens (Figure~\ref{glm_pos_compare}a).
Under ISM (Figure~\ref{glm_pos_compare}b), we retain the same 2D positional scheme, while alternating the attention pattern: bidirectional within query spans and unidirectional within answer spans. This preserves GLM’s strong span-aware positional inductive bias and maintains KV-cache reusability across turns, which achieves the contextual integration benefits without sacrificing the latency.

\begin{table}
\centering
\caption{A comparison on multi-turn dialogue benchmarks.}
\resizebox{\linewidth}{!}{
\begin{tabular}{cc|ccc|ccc|ccc}
\toprule
\multirow{2}{*}{\bf Bench} & \multirow{2}{*}{\bf Metrics} & \multicolumn{3}{c|}{\bf LLaMA3.1-8B} & \multicolumn{3}{c|}{\bf Qwen2.5-7B} & \multicolumn{3}{c}{\bf Qwen2.5-14B}\\
& & {\bf Direct} & {\bf SFT} & {\bf ISM} & {\bf Direct} & {\bf SFT} & {\bf ISM} & {\bf Direct} & {\bf SFT} & {\bf ISM}\\
\midrule

\multirow{5}{*}{\bf MT-Eval} & {\bf Expansion} & 3.66 & 6.21 & {\bf 6.80} & 4.70 & {\bf 6.63} & 6.57 & 6.43 & {\bf 7.34} & 7.03 \\
& {\bf Follow-up} & 4.23 & 7.10 & {\bf 7.15} & 6.03 & 7.71 &{\bf 7.72} & 7.55 & 8.10 & {\bf 8.37} \\
& {\bf Recollection} & 1.39 & 5.13 & {\bf 5.55} & 3.84 & 5.63 & {\bf 5.95} & 5.32 & 6.24 & {\bf 6.63}\\
& {\bf Refinement} & 2.09 & 5.03 & {\bf 5.17} & 2.92 & 5.29 & {\bf 5.36} & 4.31 & 5.68 & {\bf 5.96}\\ \cmidrule(r){2-11}
& {\bf Average} & 2.84 & 5.87 & {\bf 6.17} & 4.37 & 6.32 & {\bf 6.40} & 5.90 & 6.84 & {\bf 7.00}\\
\midrule

\multirow{2}{*}{\bf BotChat} & {\bf SR(N=16)} & 67.3\% & 77.3\% & {\bf 79.2\%} & 57.6\% & 71.8\% & {\bf 78.2\%} & 57.0\% & 85.6\% & {\bf 86.3\%} \\
& {\bf SR(N=8)} & 89.8\% & 90.5\% & {\bf 93.6\%} & 90.1\% & 89.4\% & {\bf 92.0\%} & 87.2\% & 93.4\% & {\bf 94.0\%}\\
\bottomrule
\end{tabular}}
\label{performance}
\end{table}

\section{Experiments}

\subsection{Datasets}
\begin{enumerate}
    \item \textbf{LMSYS-Chat-1M}~\cite{lmsyschat1m}. A corpus of 1M real-world conversations from 210K users. Following prior work, we organize 20K high-quality Chinese and English conversations from OpenAI and Anthropic's models with an average of 1.46 turns for training.
    \item \textbf{MuTual}~\cite{cui-etal-2020-mutual}. A multi-turn dialogue dataset derived from Chinese high school English listening comprehension exams. We use its training split of 7,363 conversations with an average of 2.14 turns.
    %\item \textbf{MATH}~\cite{mathdataset}. A collection of 12,500 competition-level mathematics problems with step-by-step solutions and boxed answers. We use the 7,500-problem training split to learn contextual understanding for problem solving.
\end{enumerate}

\subsection{Benchmarks}
\begin{enumerate}
    \item \textbf{MT-Eval}~\cite{kwan2024mteval}. A comprehensive benchmark designed to evaluate multi-turn conversational abilities. We evaluate across four interaction modes: recollection, expansion, refinement, and follow-up.
    \item \textbf{BotChat}~\cite{duan2024botchat}. An LLM-based framework for assessing multi-turn chatting capabilities. We follow the Uni-Eval protocol to calculate the Success Rate (SR) under different chat turns $N=8,16$.
    %\item \textbf{MATH}~\cite{mathdataset}. After fine-tuning on the 7,500-problem training split, we evaluate comprehensive contextual understanding and mathematical reasoning on the 5,000-problem test split.
\end{enumerate}

\subsection{Implementation Details}\label{imple}
We choose Llama3.1-8B~\cite{grattafiori2024llama3herdmodels}, Qwen2.5-7B and Qwen2.5-14B~\cite{yang2024qwen2technicalreport} as the base models and apply ISM as described in Section~\ref{practical_modify} (denoted BaseModel(ISM)). 
For each dataset, we fully fine-tune both the base model and its ISM variant for 3 epochs using 8 NVIDIA A100 GPUs.
We optimize with AdamW~\cite{loshchilov2018decoupled} at a learning rate of 1e-5 and batch size of 64.

\begin{table}
\centering
\caption{A comparison on MATH benchmarks.}
\resizebox{.6\linewidth}{!}{
\begin{tabular}{l|c}
\toprule
{\bf Methods} & {\bf Accuracy} \\
\midrule
{\bf LLaMA3.1-8B (Direct)} & 4.76\% \\
{\bf LLaMA3.1-8B (SFT)} & 15.74\% \\
\midrule
{\bf LLaMA3.1-8B (ISM)} & {\bf 18.00\%} \\
\bottomrule
\end{tabular}}
\label{math}
\end{table}

\subsection{Quantitative Analysis}

%We evaluate ISM against two baselines—vanilla model and its supervised fine-tuned (SFT) variant—on MT-Eval and BotChat as shown in Table~\ref{performance}. Overall, SFT delivers large gains over the base model, and ISM consistently improves further across all metrics, indicating that ISM’s masking complements standard fine-tuning.

We compare ISM against Direct (no task-specific tuning) and Supervised Fine-Tuned (SFT) variants on two decoder-only backbones across two benchmarks. As shown in Table~\ref{performance}:

%ISM outperforms other baselines, indicating that ISM can be applied on multiple backbones.

\begin{itemize}
    \item \textbf{MT-Eval}. ISM attains the best average score (6.17 for LLaMA3.1-8B, 6.40 for Qwen2.5-7B and 7.00 for Qwen2.5-14B), surpassing SFT and substantially improving over Direct.
    The largest gains appear in Recollection, which depends heavily on integrating prior turns, supporting that ISM enhances contextual synthesis beyond what causal mask provides.
    \item \textbf{BotChat}. 
    ISM yields consistent Success-Rate (SR) improvements under both settings: 79.2\% and 93.6\% for LLaMA3.1-8B (+1.9\% and +3.1\% over SFT), 78.2\% and 92.0\% for Qwen2.5-7B (+6.4\% and +2.6\% over SFT). 
    The stronger gains under the harder N=16 setting suggest improved multi-turn coherence and robustness.
    %\item \textbf{MATH}. Beyond dialogue, ISM improves mathematical reasoning accuracy to 18.0\%, a +2.3 pp gain over SFT and +13.2 pp over the base model. This confirms that the benefits of ISM extend to context-intensive reasoning tasks, not just conversational history modeling.
\end{itemize}
In summary, ISM delivers consistent improvements across both benchmarks and backbones. The pattern of gains indicates that ISM’s intermittent bidirectional attention effectively enhances contextual integration beyond standard SFT while remaining broadly effective across model families.

\subsection{Mathematical Analysis}
To further assess ISM on tasks requiring comprehensive contextual understanding, we choose MATH~\cite{mathdataset}, a collection of 12,500 competition-level problems with step-by-step solutions and boxed answers.
We fine-tune on the 7,500-problem training split and evaluate on the 5,000-problem test split. As shown in Table~\ref{math}, ISM improves mathematical reasoning accuracy to 18.0\%, a +2.3\% gain over SFT and +13.2\% over Direct, demonstrating that ISM’s benefits extend beyond dialogues to context-intensive mathematical reasoning.

\subsection{Latency Analysis}

% \begin{figure}
%     \centering
%     \includegraphics[width=0.48\textwidth]{figure/ttft.pdf}
%     \caption{
%     % AntGLM (ISM) reduces time-to-first-token(TTFT) by reusing KV Cache accross turns, whereas the TTFT of AntGLM with prefix mask increases rapidly as the input sequence length grows.
%     Time-to-first-token (TTFT) latency comparison between LLaMA3.1-8B(SFT) and LLaMA3.1-8B(ISM). 
%     %AntGLM (ISM) reduces TTFT significantly by reusing the KV cache across dialogue rounds.
%     }

% \begin{figure}
%     \centering
%     \includegraphics[width=0.48\textwidth]{figure/TPOT.pdf}
%     \caption{
%     % AntGLM (ISM) reduces time-to-first-token(TTFT) by reusing KV Cache accross turns, whereas the TTFT of AntGLM with prefix mask increases rapidly as the input sequence length grows.
%     TPOT latency comparison between LLaMA3.1-8B(SFT) and LLaMA3.1-8B(ISM). 
%     %AntGLM (ISM) reduces TTFT significantly by reusing the KV cache across dialogue rounds.
%     }
%     \label{latency_comparev2}
% \end{figure}

\begin{figure}
\begin{minipage}[b]{.49\linewidth}
  \centering
  \centerline{\includegraphics[width=4.8cm]{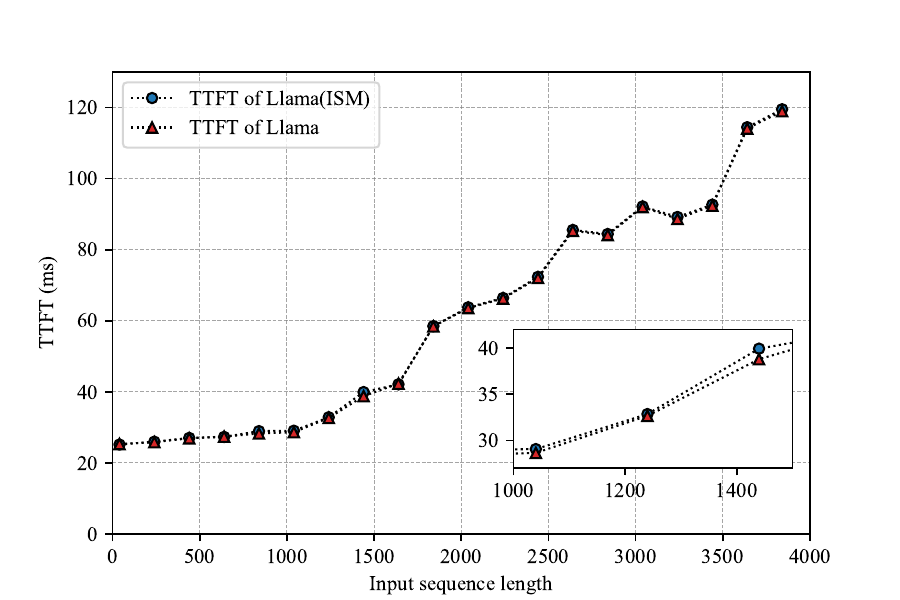}}
%  \vspace{1.5cm}
  \centerline{(a) TTFT latency}\medskip
\end{minipage}
\hfill
\begin{minipage}[b]{0.48\linewidth}
  \centering
  \centerline{\includegraphics[width=4.8cm]{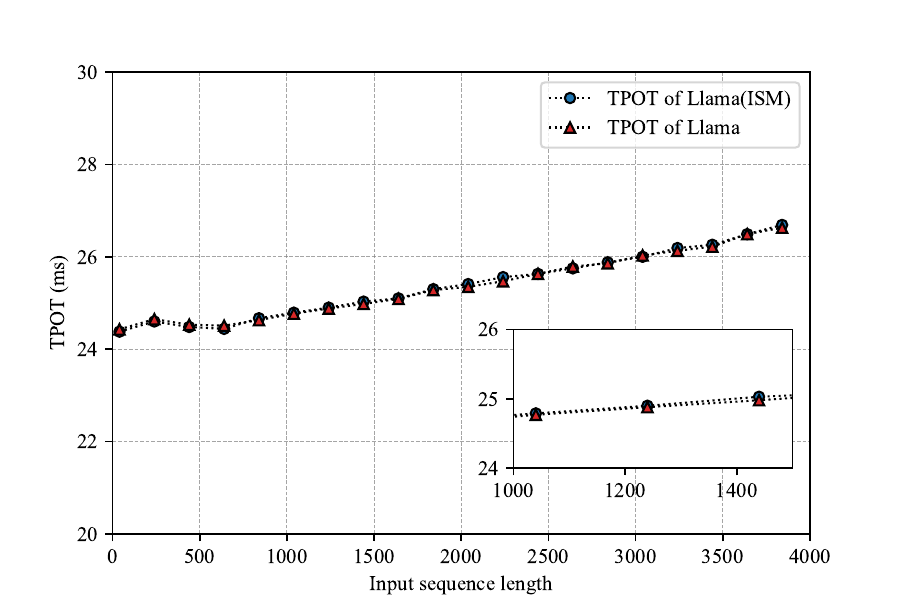}}
%  \vspace{1.5cm}
  \centerline{(b) TPOT latency}\medskip
\end{minipage}
\caption{The TTFT and TPOT latency comparison between LLaMA3.1-8B(SFT) and LLaMA3.1-8B(ISM).}
\label{latency_compare}
\end{figure}

In this section, we compare Time-To-First-Token (TTFT) and Time-Per-Output-Token (TPOT) between LLaMA3.1-8B(SFT) and LLaMA3.1-8B(ISM).
Specifically, we construct forty simulated dialogues, each with 20 turns; every turn contains 200 tokens (query length 40, answer length 160), so the effective context grows up to 4,000 tokens.
Models are deployed on a single NVIDIA A100 GPU. For each configuration, we invoke the inference API five times and report the average latency.
As shown in Figure~\ref{latency_compare}, both TTFT and TPOT scale with context length, while ISM closely tracks the SFT baseline across the entire range up to 4,000 tokens. The observed gap is small (approximately 0.5–1.0 ms), indicating near-identical latency.
These results confirm that ISM doesn't sacrifice the latency characteristics of causal mask.

\subsection{Case Study}

\begin{figure}
        \centering
        \includegraphics[width=1\linewidth]{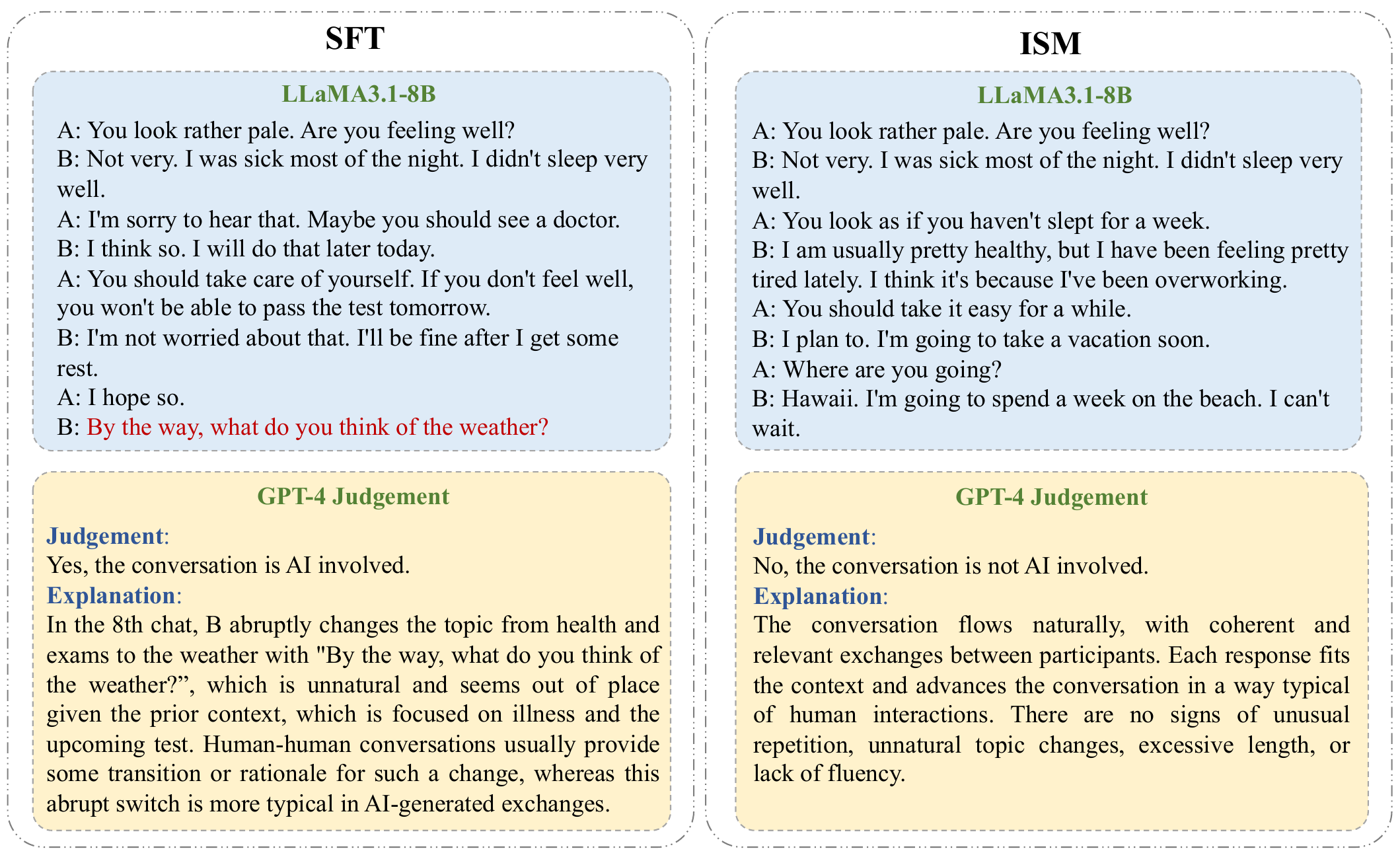}
        \caption{Two cases generated from LLaMA3.1-8B(SFT) and LLaMA3.1-8B(ISM), with GPT-4 judgements.}
        \label{casestudy}
\end{figure}

In Figure~\ref{casestudy}, We select a BotChat~\cite{duan2024botchat} benchmark sample.
The first two turns are human-written, after which each model (LLaMA3.1-8B (SFT) and LLaMA3.1-8B (ISM)) continues the conversation for several turns. Then a GPT-4 judge assesses whether the completed dialogue appears AI-involved.

The SFT model produces an abrupt, unmotivated topic shift at turn 8, breaking the ongoing thread and leading the judge to flag the conversation as AI-involved.
In contrast, the ISM model maintains topical continuity and pragmatic flow.
It illustrates two common failure modes of causal mask in multi-turn dialogue: off-topic drift and abrupt discourse shifts, and shows how ISM mitigates them.
By enabling intermittent bidirectional attention within query spans, ISM better aggregates relevant cues from the dialogue history, sustaining coherence and producing more human-like transitions.

%we showcase two cases including dialogue sample, generated responses and GPT-4 judgement results. We can find that GPT-4 follows our instructions well to make judgements. In the first case, GPT-4 not only considers the dialogue history, but also takes account of User 2's "willingness" and "responsibility". Cases with longer dialogue history can be seen in Appendix \ref{append_cases}.

%\subsection{Limitation}\label{limi}
%Current ISM is applied to existing causal or prefix base models for finetuning. The data utilized for finetuning is relatively small compared with the data used during their pretraining stage. The performance of our ISM in the pretraining stage needs to be explored.

%\section{Conclusion and Future Work}
\section{Conclusion and Future Works}
We introduced Intermittent Semi-working Mask (ISM), a simple but effective masking scheme that inserts bidirectional attention within query segments while keeping causal attention in answer segments.
This design preserves the KV-cache reuse of causal mask, yet recovers the stronger contextual integration of prefix mask.
Across MT-Eval, BotChat, and MATH, ISM improves generation quality and maintains low latency, with gains generalizing across two famous model backbones (LLaMA and Qwen).
%In this paper, we propose a novel attention masking scheme called Intermittent Semi-working Mask (ISM)
%to achieve the higher generation quality on context-intensive tasks. 
%Specifically, we apply alternate bidirectional and unidirectional attention on queries and answers in the dialogue history.
%In this way, our ISM is able to maintain the advantages of both prefix LLM and causal LLM simultaneously.
%Extensive experiments illustrate that our ISM achieves significant performance. 

In future works, we will try to (i) apply ISM during pretraining to delve deeper into exploring its advantages; (ii) extend ISM to longer-context and retrieval-augmented settings, as well as other modalities; (iii) integrate ISM with advanced inference techniques (e.g., FlashAttention, speculative decoding) while evaluating robustness, safety, and multilingual generalization in real-world deployments.

\vfill\pagebreak

\bibliographystyle{IEEEbib}
\bibliography{refs}

\end{document}